\documentclass[11pt]{article}
\usepackage{jmlr2e-mod}
\pdfoutput=1

\usepackage{graphicx}
\usepackage{booktabs}

\title{Multimodal Deep Learning for\\Flaw Detection in Software Programs}

\author{\name Scott Heidbrink \email sheidbr@sandia.gov\\
	\name Kathryn N. Rodhouse \email knrodho@sandia.gov\\
	\name Daniel M. Dunlavy\footnotemark[2] \email dmdunla@sandia.gov\\
	\addr Sandia National Laboratories\\
	\addr Albuquerque, NM 87123, USA}

\begin{document}
	
	
	\maketitle

	\renewcommand*{\thefootnote}{(\fnsymbol{footnote})}
	\footnotetext[2]{Corresponding author.}
	\renewcommand*{\thefootnote}{\arabic{footnote}.}
	
\begin{abstract}
We explore the use of multiple deep learning models for detecting flaws in software programs. Current, standard approaches for flaw detection rely on a single representation of a software program (e.g., source code or a program binary). We illustrate that, by using techniques from multimodal deep learning, we can simultaneously leverage multiple representations of software programs to improve flaw detection over single representation analyses. Specifically, we adapt three deep learning models from the multimodal learning literature for use in flaw detection and demonstrate how these models outperform traditional deep learning models. We present results on detecting software flaws using the Juliet Test Suite and Linux Kernel.

\end{abstract}


\begin{figure}[b!]
\centering
\includegraphics[width=\textwidth]{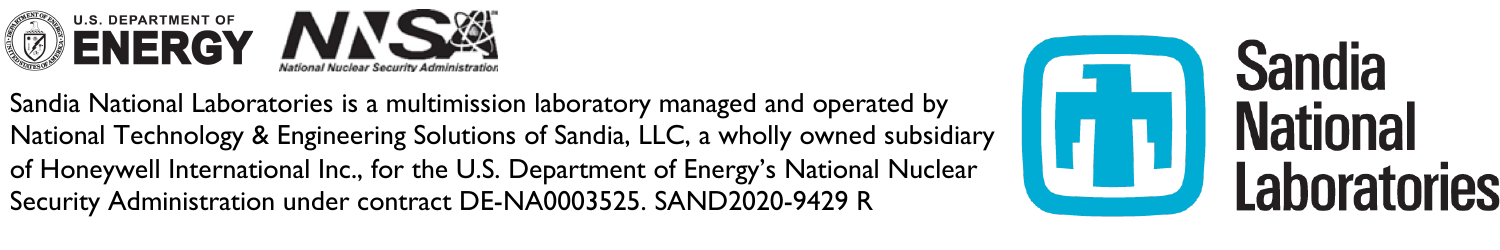}
\end{figure}

\section{Introduction}
\label{sec:intro}
Efficient, reliable, hardened software plays a critical role in cybersecurity.  Auditing software for flaws is still largely a manual process.  Manual review of software becomes increasingly intractable as software grows in size and is constantly updated through increasingly rapid version releases.  Thus, there is a need to automate the identification of flawed code where possible, enabling auditors to make the most efficient use of their scant time and attention.

Most current approaches for flaw detection rely on analysis of a single representation of a software program (e.g., source code or program binary compiled in a specific way for a specific hardware architecture).  This is often the case when analyzing commercial software or embedded system software developed by external groups, where program binaries are available without associated source code or build environments.  However, different program representations can 1) contain unique information, and 2) provide specific information more conveniently than others.  For instance, source code may include variable names and comments that are removed through compilation, and program binaries may omit irrelevant code detected through compiler optimization processes. Thus, there can be an added benefit in understanding software using multiple program representations.  

In this paper, we explore the use of multiple program representations (i.e., source code and program binary) to develop machine learning models capable of flaw detection, even when only a single representation is available for analysis (i.e., a program binary). Our contributions include:
\begin{itemize}
	\item The first application of multimodal learning for software flaw prediction (as far as we are aware);
	\item A comparative study of three deep learning architectures for multimodal and learning applied to software flaw prediction; and
    \item A data set of software flaws with alignment across source code and binary function instances that can be used by the multimodal learning and software analysis research communities for benchmarking new methods. 
\end{itemize}

The remainder of this paper is organized as follows. We start with a discussion on related work in Section~\ref{sec:related}. In Section~\ref{sec:data}, we describe the software flaw data sets that we use in  our comparative studies. In Section~\ref{sec:methods}, we describe the deep learning architectures that we use  and the experiments that compare these different architectures for flaw detection, and in Section~\ref{sec:results}, we report on the results of those experiments. Finally, in Section~\ref{sec:conc}, we provide a summary of our findings and suggest several paths forward for this research.

\section{Related Work}
\label{sec:related}
Multimodal learning is a general approach for relating multiple representations of data instances. For example, in speech recognition applications, Ngiam \emph{et al.} demonstrated that audio and video inputs could be used in a multimodal learning framework to 1) create improved speech classifiers over using single modality inputs, and 2) reconstruct input modalities that were missing at model evaluation time (also known as crossmodal learning) \cite{NgKhKi11}. 

Much of the work in multimodal and crossmodal learning has focused on recognition and prediction problems associated with two or more representations of transcript, audio, video, or image data \cite{ChKhLaRa16,NgKhKi11,VuRaGr16,EpMeMi18,WaArLi15,SoGaMaNg13}. We aim to leverage those methods and results to develop an improved software flaw detector that learns from both source code and binary representations of software programs.

Recent research has demonstrated the utility of using deep learning to identify flaws in source code by using statistical features captured from source, sequences of source code tokens, and program graphs \cite{GhSh2017,AlBaDeSu2017}.  Other research efforts have demonstrated the utility of using deep learning to identify flaws in software binaries \cite{XuSuVeLa2019,LeKwChLi19,TiXiLi20,LiZoXuCh20,ArHaKlGa20}. In all of this previous work, there is clear evidence that using deep learning over traditional machine learning methods helps improve flaw prediction. Furthermore, more recently, Harer, {\it et al.}~\cite{HaKiRu18} demonstrated improved flaw modeling by combining information from both the source and binary representations of code. Although not presented as such, the latter work can be considered an instance of multimodal learning using early fusion. Our work differs from Harer, {\it et al.}~\cite{HaKiRu18} in that we employ models that learn joint representations of the input modalities, which have been demonstrated to outperform early fusion models for many applications~\cite{ChKhLaRa16,NgKhKi11,VuRaGr16,EpMeMi18,WaArLi15,SoGaMaNg13}.

Gibert {\it et al.}~\cite{GiMaPl20} describe recent advances in malware detection and classification in a survey covering multimodal learning approaches. They argue that deep learning and multimodal learning approaches provide several benefits over traditional machine learning methods for such problems. Although the survey covers similar multimodal learning methods to those we investigate here (i.e., {\it intermediate fusion methods}),  those methods for malware analysis leverage sequences of assembly instructions and Portable Execution metadata/import information as features, whereas in our work we leverage source code and static binary features.

\section{Data}
\label{sec:data}
We use two data sets, described below, to assess the performance of our methods in detecting flaws. Flaws are labeled in source code at the function level; i.e., if one or more flaws appear in a function, we label that function as {\it flawed}, otherwise it is labeled as {\it not flawed}. We compile all source code in these data sets using the GCC 7.4.0 compiler [{gcc.gnu.org}] with debug flags, which enables mapping of functions between source code and binaries. 
Since we focus on multimodal deep learning in this work, we use only the functions that we can map one-to-one between the source code and binary representations in our experiments.

\subsection{Juliet Test Suite}

The Juliet Test Suite~\cite{NIST}, which is part of NIST's Software Assurance Reference Database, encompasses a collection of C/C++ language test cases that demonstrate common software flaws, categorized by Common Weakness Enumeration (CWE) [{cwe.mitre.org}].  Previous research efforts using machine learning to identify flaws in software have used this test suite for benchmark assessments~\cite{HaOzLa18,RuKiHa18,LeKwChLi19,TiXiLi20,LiZoXuCh20,ArHaKlGa20}. We use a subset of test cases covering a variety of CWEs (both in terms of flaw type and number of functions available for training models) to assess method generalization in detecting multiple types of software flaws. The labels of {\it bad} and {\it good} defined per function in the Juliet Test Suite map to our labels of {\it flawed} and {\it not flawed}, respectively. Table~\ref{tab:juliet} presents information on the specific subset we use, including a description and size of each CWE category.

\begin{table}[ht!]
	\centering

	\begin{tabular}{cll}
		\toprule
		& & \# Flawed;\\
		CWE & Flaw Description & \# Not Flawed\\
		\midrule
121 & Stack Based Buffer Overflow & 6346; 16868 \\
190 & Integer Overflow                 & 3296; 12422  \\
369 & Divide by Zero                    & 1100; 4142  \\
377 & Insecure Temporary File      & 146; 554  \\
416 & Use After Free                    & 152; 779  \\
476 & NULL Pointer Dereference    & 398; 1517  \\
590 & Free Memory Not on Heap   & 956; 2450  \\
680 & Integer to Buffer Overflow   & 368; 938  \\
789 & Uncontrolled Mem Alloc       & 612; 2302  \\
78   & OS Command Injection       & 6102; 15602  \\ 
\bottomrule
	\end{tabular}
		\caption{Juliet Test Suite Data Summary}
	\label{tab:juliet}
\end{table}

\subsection{Flaw-Injected Linux Kernel}

The National Vulnerability Database [{nvd.nist.gov}], a repository of standards based vulnerability management, includes entries CVE-2008-5134 and CVE-2008-5025, two buffer overflow vulnerabilities associated with improper use of the function \texttt{memcpy} within the Linux Kernel [{www.kernel.org}]. We develop a data set based on these vulnerabilities to demonstrate the performance of our methods on a complex, modern code base more indicative of realistic instances of flaws than are provided by the Juliet Test Suite. Specifically, we inject flaws into Linux Kernel 5.1 by corrupting the third parameter of calls to the function \texttt{memcpy}, reflecting the pattern of improper use found in the two vulnerabilities mentioned above. We note two assumptions we make in this process: 1) the original call to \texttt{memcpy} is not flawed, and 2) by changing the code we have injected a flaw.  We expect each of these assumptions to generally hold for stable software products.  
There are more than 530,000 functions in the Linux Kernel 5.1 source code. We corrupt 12,791 calls to \texttt{memcpy} in 8,273 of these functions. After compiling using the default configuration, these flaws appear in 1,011 functions identified in the binaries. 

\subsection{Source Code Features}
\label{sec:source}

We use a custom program graph extractor to generate the following structures from the C/C++ source code in our data: abstract syntax tree (AST), control flow graph (CFG), inter-procedural control flow graph, scope graph, use-def graph (UDG), and type graph.  From these graphs, we extract flaw analysis-inspired statistical features associated with the following program constructs~\cite{GhSh2017}:
\begin{itemize}
    \item called/calling functions (e.g., number of external calls)
    \item variables (e.g., number of explicitly defined variables)
    \item graph node counts (e.g., number of else statements)
    \item graph structure (e.g., degrees of AST nodes by type)
\end{itemize}

In addition to these statistical features, we also capture subgraph/subtree information by counting all unique node---edge---node transitions for each of the these graphs following Yamaguchi {\it et al.}~\cite{YaLoRi12}, which demonstrated the utility of such features in identifying vulnerabilities in source code. 
Examples include:
\begin{itemize}
\item AST:CallExpression---argument---BinaryOperator
\item CFG:BreakStatement---next---ContinueStatement
\item UDG:MemberDeclaration---declaration---MemberDeclaration
\end{itemize}  

Following this procedure, we extracted 722 features for the Juliet Test Suite functions and 1,744 features for the Linux Kernel functions. 

\subsection{Static Binary Analysis Features}
\label{sec:binary}
We use the Ghidra 9.1.2 software reverse engineering tool [{ghidra-sre.org}] to extract features from binaries following suggestions in Eschweiler {\it et al.}~\cite{EsYaGe16}, which demonstrated the utility of such features in identifying security vulnerabilities in binary code. Specifically, we collect statistical count information per function associated with the following:
\begin{itemize}
    \item called/calling functions (e.g., number of call out sites)
    \item variables (e.g., number of stack variables)
    \item function size (e.g., number of basic blocks)
    \item p-code\footnote{{\it p-code} is Ghidra's intermediate representation/intermediate language (IR/IL) for assembly language instructions} opcode instances (e.g., number of COPYs)
\end{itemize}

Following this procedure,  we extracted 77 features for both the Juliet Test Suite and Linux Kernel functions.

\section{Methods}
\label{sec:methods}
\subsection{Multimodal Deep Learning Models}
\begin{figure*}[!ht]
	\centering
	\includegraphics[width=\linewidth]{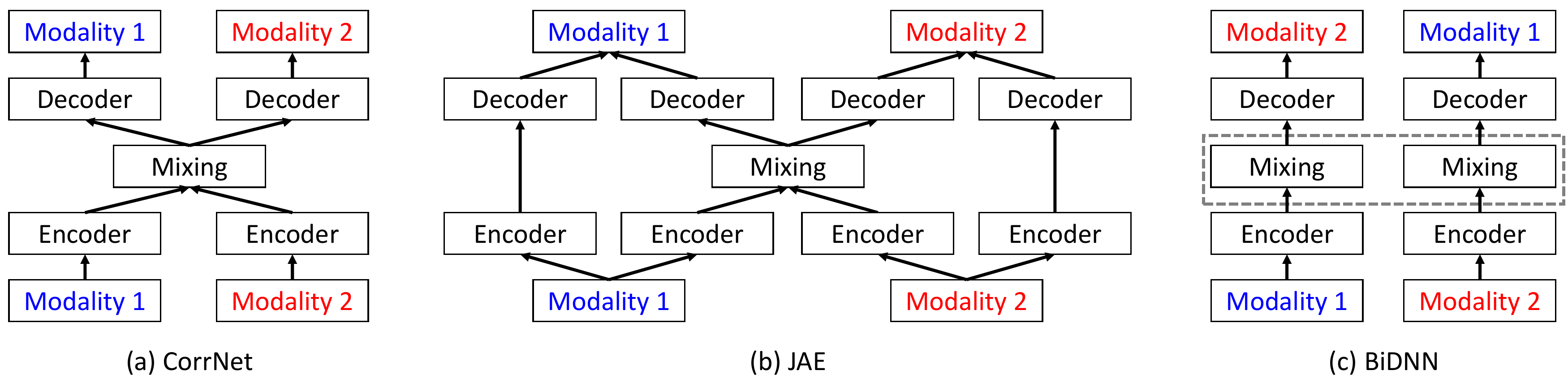}
    \caption{The general architectures for the three approaches examined in this work: (a) CorrNet, (b) JAE, and (c) BiDNN.}
	\label{fig:arch}
\end{figure*}

We investigate three multimodal deep learning architectures for detecting flaws in software programs. Figure~\ref{fig:arch} presents high level schematics of the different architectures, highlighting the main differences between them. These models represent both the current diversity of architecture types in joint representation multimodal learning as well as the best performing models across those types~\cite{BaAhMo19,WaArLi15,EpMeMi18,VuRaGr16}. 

\subsubsection{Correlation Neural Network (CorrNet)}

The Correlation Neural Network (CorrNet) architecture is an autoencoder containing two or more inputs/outputs and a loss function term that maximizes correlation across the different input/output modalities~\cite{ChKhLaRa16}. Figure~\ref{fig:arch}(a) illustrates the general architecture of CorrNet, where each of the Encoder, Mixing, and Decoder layers are variable in both the number of network layers and number of nodes per layer. Wang \emph{et al.} extended CorrNet to use a deep autoencoder architecture that has shown promise on several applications~\cite{WaArLi15}. The main distinguishing feature of CorrNet models is the use of a loss function term taken from Canonical Correlation Analysis (CCA) models, where the goal is to find a reduced-dimension vector space in which linear projections of two data sets are maximally correlated~\cite{Ho36}. The contribution of this CCA loss term used in CorrNet is weighted, using a scalar term denoted as $\lambda$, to balance the impact of CCA loss with other autoencoder loss function terms. The implementation of the CorrNet model used in our experiments follows Wang, {\it et al.}~\cite{WaArLi15}.

\subsubsection{Joint Autoencoder (JAE)}
The Joint Autoencoder (JAE) model was originally developed as a unified framework for various types of meta-learning (e.g., multi-task learning, transfer learning, multimodal learning, etc.)~\cite{EpMeMi18}. JAE models include additional Encoder and Decoder layers for each of the modalities---denoted as private branches---that do not contain mixing layers. Figure~\ref{fig:arch}(b) illustrates the general architecture of the JAE model. The additional private branches provide a mechanism for balancing contributions from each modality separately and contributions from the crossmodal Mixing layers. Each of the Encoder, Mixing, and Decoder layers of the JAE model are variable in both the number of network layers and number of nodes per layer. 

\subsubsection{Bidirectional Deep Neural Network (BiDNN)}
The Bidirectional Deep Neural Network (BiDNN) model performs multimodal representational learning using two separate neural networks to translate one modality to the other~\cite{VuRaGr16}.
The weights associated with the Mixing layer are symmetrically tied, as denoted in Figure~\ref{fig:arch}(c) by the dotted lines around that layer for each of the networks. Since the Mixing layer weights are tied across the modalities, the Mixing layers provide a single, shared representation for multiple modalities. Each of the Encoder, Mixing, and Decoder layers of the BiDNN model are variable in both the number of network layers and number of nodes per layer.

\subsection{Experimental Setup}
\label{sec:experimental_setup}
Experiments use the three deep learning architectures---CorrNet, JAE, and BiDNN---presented in Section~\ref{sec:methods} and the data presented in Section~\ref{sec:data}. The goals of these experiments are to answer the following questions:
\begin{itemize}
\item Does using multimodal deep learning models that leverage multiple representations of software programs help improve automated flaw prediction over single representation models?
\item How sensitive are the flaw prediction results of these models as a function of the architecture choices and model parameters?
\item Are there differences in flaw prediction performance across the various deep learning models explored?
\end{itemize} 

The data is first normalized, per feature, to have sample mean of 0 and sample standard deviation of 1, and split into standard training sizes of (80\%), validation (10\%), and testing sets (10\%).
In the cases where a feature across all training set instances is constant, the sample standard deviation is set to 0 so that the feature does not adversely impact the network weight assignments in the validation and testing phases.
We use 5-fold cross validation to fit and evaluate instances of each model to assess how well our approach  generalizes. 

We implement each architecture in PyTorch v1.5.1 using Linear layers and LeakyReLU activations on each layer excluding the final one. In all experiments, we use PyTorch's default parameterized Adam optimizer, and, excluding the Initialization experiment, PyTorch's default Kaiming initialization with LeakyReLU gain adjustment, and 100 epochs for training. To regularize the models, we use the best performing parameters as evaluated on the fold's validation set. In all experiments but the Architecture Size experiment, each of the Encoder, Mixing, and Decoder layers contain $50$ nodes and $1$ layer. For all decisions not explicitly stated or varied within the experiment, we rely on PyTorch's default behavior.

We construct a deep neural network with the same number of parameters as the multimodal deep learning models and use it as the {\it baseline classifier} in our experiments. These baseline classifier models are composed of the Encoder and Mixing layers (see Figure~\ref{fig:arch}) followed by two Linear layers. This approach is an instance of early fusion multimodal deep learning as demonstrated in \cite{HaKiRu18} as an improvement over single modality deep learning models for flaw prediction. 

Our experiments consist of predicting flaws on several Juliet Test Suite CWE test cases and the flaw-injected Linux Kernel data, both described in Section~\ref{sec:data}. We also perform several experiments using the multimodal deep learning models on the flaw-injected Linux Kernel data to study the impact of the construction and training of these models on the performance of predicting flaws. Specifically, we assess the impacts of 1) the size and shape of the neural networks, 2) the method used for setting initial neural network weights, 3) using both single and multimodal inputs versus multimodal inputs alone, and 4) the amount of correlation that is used in training the CorrNet models. We include the results of these additional experiments to provide insight into the robustness of the multimodal deep learning models in predicting flaws.

\section{Results}
\label{sec:results}
In this section we present the results of our experiments of using multimodal deep learning models for flaw prediction. 
In stable software products, the number of functions containing flaws is much smaller than the number of functions that do not contain flaws. This class imbalance of {\it flawed} and {\it not flawed} functions is also reflected in the Juliet Test Suite and flaw-injected Linux Kernel data that we use for our experiments. Thus, when reporting model performance, we report accuracy weighted by the inverse of the size of the class to control for any bias.

\begin{figure*}[!ht]
	\centering
	\includegraphics[width=\linewidth]{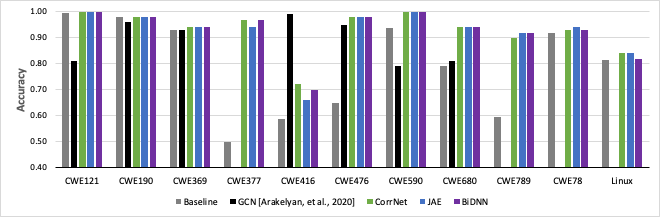}
    \caption{Flaw Detection Results on Juliet Test Suite (CWE) and Linux Kernel Data Sets}
	\label{fig:flaw_detection}
\end{figure*}

\subsection{Flaw Detection Results}
\label{sec:baseline_results}
In this section we present the overall results of our experiments of using multimodal deep learning models for flaw prediction. Figure~\ref{fig:flaw_detection} presents the best results across all experiments performed involving the baseline, CorrNet, JAE, and BiDNN models on the Juliet Test Suite (denoted by CWE) and the flaw-injected Linux Kernel data (denoted by Linux). The values in the figure reflect the averages of accuracy across the 5-fold cross validation experiments. As noted in Section~\ref{sec:data}, there are numerous published results on using the Juliet Test Suite in assessing deep learning approaches to flaw predictions. However, existing results cover a wide variety of flaws across the various CWE categories, and there is no single set of CWEs that are used in all assessments. To illustrate comparison of the multimodal deep learning models to published results, we also include in Figure~\ref{fig:flaw_detection} recent results of using a graph convolutional network (GCN) deep learning model for flaw prediction~\cite{ArHaKlGa20} for the CWEs which overlap between our results and theirs. 

In all experiments, multimodal deep learning models perform significantly better than the baseline deep learning models.
Moreover, in all but one experiment, the multimodal deep learning models perform better (and often significantly better) than the published results for the GCN deep learning models. In that one exception, CWE416, the multimodal deep learning models perform the worst across data explored in our research presented here. We hypothesize that the diminished performance of our multimodal methods is partly due to the small size of the data---there are fewer than 1000 total functions in CWE416---as deep learning models often require a lot of training data for good performance. However, it could also be due to the specific type of flaw (Use After Free) in that category. Determining the sources of such differences between published results and the results we present here is left as future work.

Compared with other published results on predicting flaws in the Juliet Test Suite, the multimodal deep learning models perform as well as (and often better than) existing machine learning approaches. Although not as many specific, direct comparisons can be made as with the GCN results discussed above, we present here several comparisons with published results. Instruction2vec~\cite{LeKwChLi19} uses convolutional neural networks on assembly instruction input to achieve ~0.97 accuracy on CWE121 compared to the multimodal deep learning model results of greater than 0.99 accuracy shown in Figure~\ref{fig:flaw_detection}. VulDeeLocator~\cite{LiZoXuCh20} uses bidirectional recurrent neural networks on source code and intermediate representations from the LLVM compiler to achieve accuracies of 0.77 and 0.97, respectively, across a collection of Juliet Test Suite CWE categories; this is comparable to our results where on average across all Juliet CWEs tested the multimodal deep learning models achieve an average accuracy of 0.95 (leaving out the anomalous results for CWE416 as discussed above). And BVDetector~\cite{TiXiLi20} uses bidirectional neural networks on graph features from binaries to achieve at most 0.91 accuracy across collections of the Juliet Test Suite associated with memory corruption and numerical issues, where the multimodal deep learning models on average achieve 0.96 accuracy on data related to those flaw types (i.e., CWE121, CWE190, CWE369, CWE590, CWE680, and CWE789). 

\subsection{Multimodal Model Parameterization Results}
\label{sec:results_multimodal}

Each of the three different deep learning models we explore---CorrNet, JAE, and BiDNN---can be constructed, parameterized, and trained in a variety of ways. In this section, we present some of the most notable choices and analyze their impact on reconstruction and classification performance for software flaw modeling. We present here only the results on flaw-injected Linux Kernel data in an attempt to reflect potential behaviors of using these models on modern, complex software programs. Performance on the Juliet Test Suite is comparable and supports similar conclusions.

\subsubsection{Architecture Size}
\label{sec:results:size}
We first examine the impact of the architecture size (i.e., the model depth or number of layers, the nodes per layer, and overall model parameters) by comparing model instances of each architecture that have the same number of overall model parameters. JAE models have two additional private branch Encoder and Decoder layers, which effectively double the size of the model compared to the CorrNet and BiDNN architecture models; thus, we only use half the number of nodes per layer for each JAE model.  In these experiments, we set the value of $\lambda$ in the CorrNet models to 0, effectively removing the correlation loss term. We explore the effect of the CorrNet $\lambda$ value later in this section.

Table~\ref{tab:sizeflaw} illustrates the effect of architecture size on model performance.  The layer size and depth refer to each of the Encoder, Mixing, and Decoder layers of the models (see Figure~\ref{fig:arch}). Results are presented as averages across the cross validation results. The best results per architecture per performance measure are highlighted in bold text. As is the case with many deep learning models, the general trends of the results indicate that deeper networks performs best. Overall, BiDNN performs the best in terms of flaw classification, but the differences in performance are not significant.

\begin{table}[h!]
	\centering
	\begin{tabular}{crrr}
		\toprule
		(layer size $\times$ layer depth) &       CorrNet &           JAE &         BiDNN \\
		\midrule
		          $50 \times 1$           &          0.78 &          0.80 &          0.79 \\
		         $100 \times 1$           &          0.76 &          0.78 &          0.76 \\
		         $500 \times 1$           &          0.79 &          0.79 & \textbf{0.82} \\
		         $100 \times 2$           &          0.80 & \textbf{0.83} &          0.81 \\
		          $50 \times 4$           & \textbf{0.81} &          0.80 &          0.80 \\ 
		          \bottomrule
	\end{tabular}
		\caption{Architecture Size Results}
	\label{tab:sizeflaw}
\end{table}

\subsubsection{Model Weights Initialization}
\label{sec:model_weights}
The weights of the deep learning models explored in this work can be initialized using a variety of approaches. For example, the authors of the original JAE and BiDNN models use Xavier initialization~\cite{glorot2010understanding}. He \emph{et al.} demonstrate that when using LeakyReLU activation, Kaiming initialization may lead to improved model performance~\cite{he2015delving}. Furthermore, LSUV initialization has recently gained popularity~\cite{mishkin2015all}.
In this section, we present the impact of weight initialization schemes on model performance, baselining with constant initialization; bias values are set using random initialization.

Table~\ref{tab:init} presents the results of using the different initialization methods with the different multimodal deep learning models. As in the previous experiments, the best model performances per architecture per performance measure are highlighted in bold. From these results, we see that there is not significant variation across the models using the different types of initialization. However, since the best initialization method varies across the different models, we recommend comparing these methods when applying multimodal deep learning models in practice.

\begin{table}[ht!]
	\centering
	\begin{tabular}{crrr}
		\toprule
		 Initialization   &       CorrNet &           JAE &         BiDNN \\ \midrule
		    Constant      &          0.76 &  \textbf{0.84}&          0.80 \\
		     Kaiming      & \textbf{0.80} &          0.80 & \textbf{0.81} \\
		     Xavier       &          0.78 &          0.78 &          0.78 \\
		      LSUV        &          0.79 &          0.79 &          0.80 \\ \bottomrule
	\end{tabular}
	\caption{Model Weights Initialization Results}
	\label{tab:init}
\end{table}

\subsubsection{CorrNet Correlation Parameterization}
\label{sec:model_param}
The only unique parameter across the deep learning models explored in this work is CorrNet's $\lambda$ value, which balances the correlation loss term with the autoencoder loss terms.
We vary the values of $\lambda$ to determine its effect on model performance; results are shown in Table~\ref{tab:params}.
We include several $\lambda$ values in the range of $[0,10]$ and an empirical $\lambda$ value (``auto'') that equalizes the magnitudes of the correlation loss term with the autoencoder terms for a sample of the training data (as recommended by Chandar, {\it et al.}~\cite{ChKhLaRa16}). 
A small correlation ($\lambda = .1$) performs best, it seems there is no significant difference in including it as a loss term as long as it is not unduly weighted.

\begin{table}[ht!]
	\centering
	\begin{tabular}{cccccc}
		\toprule
		$\lambda$ =   0 & $\lambda$ =0.01 & $\lambda$ =0.1 & $\lambda$ =1 & $\lambda$ =10 & auto   $\lambda$ \\ \midrule
		0.78 &      0.80       &  \textbf{0.80}      &    0.74     &     0.72      &       0.79       \\ \bottomrule
	\end{tabular}
	\caption{CorrNet Correlation Parameterization Results}
	\label{tab:params}
\end{table}

\subsubsection{Single Multimodal Inputs}
\label{sec:singlemodal}
The original CorrNet and BiDNN authors recommend training the models using only single modality inputs, supplying zero vectors for the other modality, to help improve model robustness.
The CorrNet model includes this behavior explicitly within its loss function, but the JAE and BiDNN models do not contain loss function terms to account for this explicitly.
In this experiment, we evaluate the impact of using a combination of single and multimodal inputs for training.  We first augment our dataset by adding instances that have one modality zeroed out, resulting in a dataset with three times as many instances as the original. 
Otherwise, model training is conducted similarly to the previous experiments.

As shown in Table~\ref{tab:single}, the addition of single modality training data hinders model performance. This is a surprising result, especially given the current recommendations in the multimodal deep learning literature. 
Furthermore, as single modality training is part of the default CorrNet and BiDNN models, we suggest future research focus on this potential discrepancy when applying to flaw prediction problems.

\begin{table}[ht!]
	\centering
	\begin{tabular}{lrrr}
		\toprule
		Model Inputs      & CorrNet &  JAE & BiDNN \\ \midrule
		Single+Multimodal &    0.74 & 0.74 &  0.72 \\
		Multimodal        &    0.78 & 0.80 &  0.79 \\ \bottomrule
	\end{tabular}
	\caption{Single+Multimodal vs. Multimodal Results}
	\label{tab:single}
\end{table}

\section{Conclusions}
\label{sec:conc}
As discussed in Section~\ref{sec:experimental_setup}, we set out to answer the following questions: 1) can multimodal models improve flaw prediction accuracy, 2) how sensitive is the performance of multimodal models with respect to model parameter choices as applied to software flaw prediction, and 3) does one of the evaluated multimodal models outperform the others?

In Section~\ref{sec:baseline_results} we demonstrated that the multimodal deep learning models---CorrNet, JAE, and BiDNN---can significantly improve performance over other deep learning methods in predicting flaws in software programs. In Section~\ref{sec:results_multimodal}, we addressed the second question of parameter sensitivities associated with multimodal deep learning models, illustrating the relative robustness of these methods across various model sizes, model initializations, and model training approaches. We see across all of the results presented in Section~\ref{sec:results} that amongst the three multimodal deep learning models we studied in this work, no one model is clearly better than the others in predicting flaws in software programs across all flaw types. Deeper examination of the individual flaw predictions could provide a better understanding of the differences between these three models and identify which model may be best for different flaw types encountered by auditors.

In the case where sufficient training data is available, performance of the multimodal deep learning models was much higher on the Juliet Test Suite compared to that of the flaw-injected Linux Kernel data (see Figure~\ref{fig:flaw_detection}). A main difference between these data sets (as described in Section~\ref{sec:data}), is that the Linux Kernel data is comprised of functions from a modern, complex code base, whereas the Juliet Test Suite is comprised of independent examples of flaws designed to enumerate various contexts in which those flaws may arise within single functions. We believe the Linux Kernel data, inspired by real flaws logged within the National Vulnerability Database, reflects more realistic software flaws, and for this reason, presents more difficult flaw prediction problems than the Juliet Test Suite. Thus, we share this data set for use by the machine learning and software analysis communities as a potential benchmark. However, this data is limited to a single flaw type, and software programs often contain several different types of flaws concurrently. Future work should address this by providing larger data sets with sufficient complexity and flaw type variability.




\end{document}